\documentclass[letterpaper, 10 pt, conference]{ieeeconf}  
\IEEEoverridecommandlockouts                             
\usepackage{graphics} 
\usepackage{epsfig} 
\usepackage{mathptmx} 
\usepackage{times} 
\usepackage{amsmath} 
\usepackage{amssymb}  
\usepackage{cite}
\usepackage{booktabs}
\usepackage{multirow}
\usepackage[dvipsnames]{xcolor}
\DeclareMathOperator*{\argmax}{argmax}
\definecolor{mygrey}{RGB}{128, 128, 128}
\definecolor{myorg}{RGB}{236, 117, 73}
\definecolor{myblue}{RGB}{143, 170, 220}

\title{\LARGE \bf
 Object-centric Inference for Language Conditioned Placement:\\A Foundation Model based Approach 
}

\author{Zhixuan Xu, Kechun Xu, Yue Wang, Rong Xiong
\thanks{Zhixuan Xu, Kechun Xu, Yue Wang, Rong Xiong are with Zhejiang University, Hangzhou, China. Corresponding author, {\tt \small wangyue@iipc.zju.edu.cn.}}
}

\begin{document}

\maketitle
\thispagestyle{empty}
\pagestyle{empty}

\begin{abstract}

We focus on the task of language-conditioned object placement, in which a robot should generate placements that satisfy all the spatial relational constraints in language instructions. Previous works based on rule-based language parsing or scene-centric visual representation have restrictions on the form of instructions and reference objects or require large amounts of training data. We propose an object-centric framework that leverages foundation models to ground the reference objects and spatial relations for placement, which is more sample efficient and generalizable. Experiments indicate that our model can achieve a 97.75\% success rate of placement with only $\sim$0.26M  trainable parameters. Besides, our method generalizes better to both unseen objects and instructions. Moreover, with only 25\% training data, we still outperform the top competing approach.

\end{abstract}

\section{Introduction}

\begin{figure}[tb] \centering
    \includegraphics[width=0.98 \linewidth]{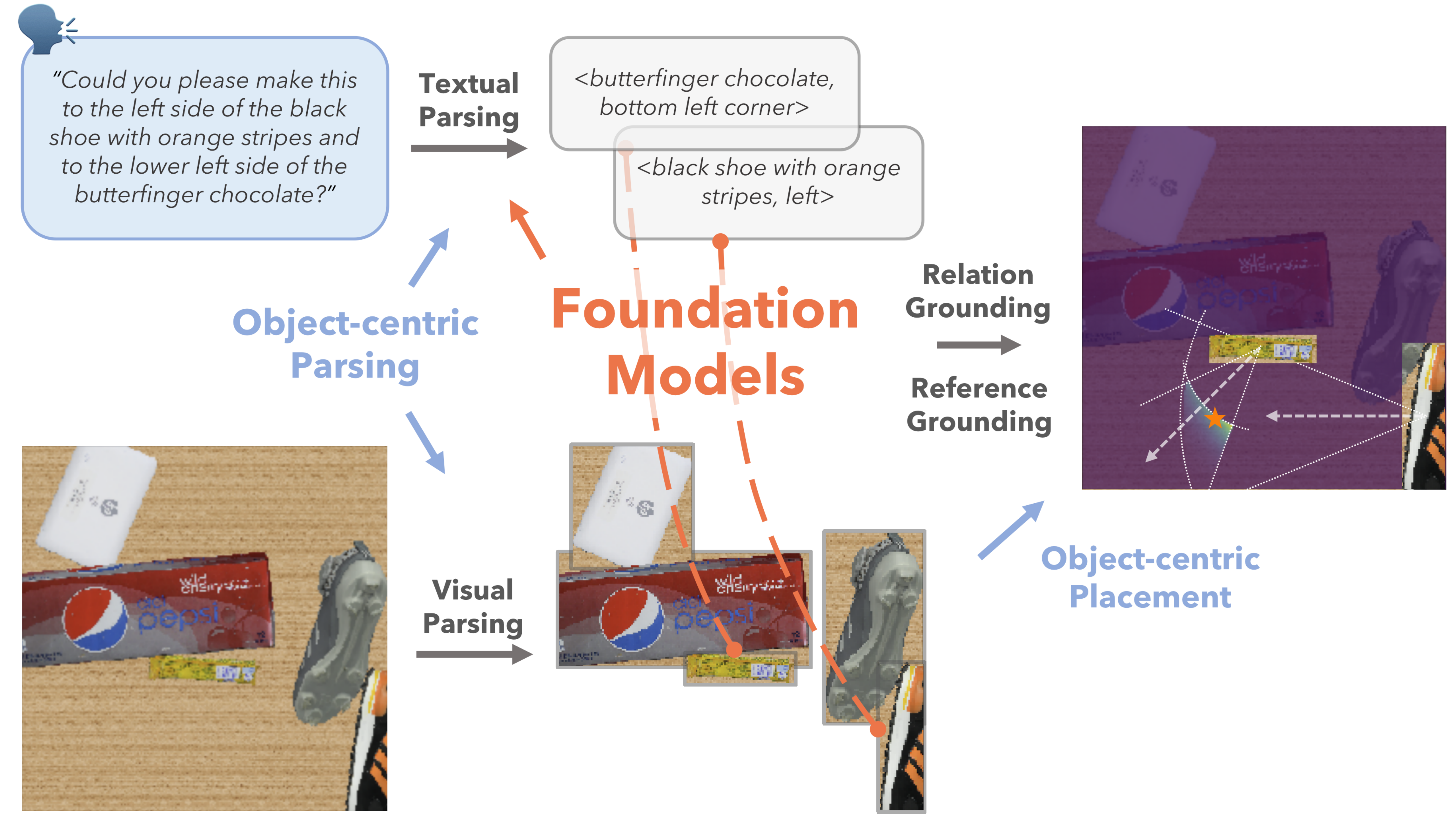}
    \vspace{-10pt}
    
    \caption{An example scenario of language-conditioned placement. We incorporate the foundation models into an object-centric framework and generate placements that satisfy all spatial relational constraints contained in instructions.} \label{fig:figure1}
    \vspace{-23pt}
\end{figure}

Object placement is an essential task in human-robot interaction. In this task, the robot should place an object in a specific location according to the goals given by the users. Previous works provide goal images \cite{gctn} or collect additional demonstrations \cite{tn}, which is unscalable and infeasible. Since objects are often described in relation to others, natural language provides an intuitive interface to specify goals. In the language-conditioned placement task, instructions are generally in the form of multiple groups of reference objects and their corresponding spatial relations, as shown in Fig. \ref{fig:figure1}. Understanding reference objects and spatial relations in instructions is therefore crucial for language-conditioned placement. 

Recently, several frameworks for language-conditioned object placement have been proposed \cite{mees1, mees2, paragon, cliport, semantic}. While these methods can generate correct object placement, they can not adapt to objects and instructions in the open world, or have low sample efficiency. From the perspective of language instruction processing, \cite{mees1, mees2, semantic} use hand-crafted rules and syntactic structures to parse instructions, which can not be generalized to flexible instructions. In \cite{mees1, mees2}, only language instructions with one reference object and spatial relation are supported. CLIPort \cite{cliport} fuses the whole instruction sentence embeddings into a convolutional architecture. However, it's not data-efficient and difficult to generalize to long or unseen instructions by directly using sentence embeddings. The latest work \cite{semantic} tries to combine rule-based and embedding-based methods, but it relies on the accuracy of the dependency tree and requires a large amount of data for training the soft parsing module. Regarding visual processing, \cite{mees1} contains only one object in the scene and does not support reference object grounding. \cite{mees2} requires training reference object grounding on a large amount of data (RefCOCO \cite{refcoco}). CLIPort \cite{cliport} is a scene-centric method that predicts pixel-wise affordance on raw images. Its fully convolutional architecture can capture strong local correlations, which helps generate ``inside'' placement, but it has worse performance when it comes to non-local correlations such as ``left to''. Paragon \cite{paragon} utilizes object-centric representation but directly uses the pre-trained CLIP \cite{clip} for reference object grounding. So it only works well when the reference objects are common and cannot transfer to more concrete and novel reference objects involved in the current task. 

Inspired by the human way of describing a placement by first anchoring on a reference object and then deciding on relative spatial relations, we propose a natural framework to generate placements by first visually grounding and anchoring to reference objects and then specifying spatial relations. Compared to previous works, our approach is more sample efficient than learning the pixel-wise placement affordance directly. For language instructions processing, we deploy the pre-trained GPT-3 \cite{gpt} and the pre-trained CLIP \cite{clip} to parse and encode instructions, allowing for handling more flexible instructions. For visual processing, we leverage the pre-trained CLIP \cite{clip} with object-centric representation, enhancing sample efficiency and relieving the sim2real problem \cite{vilg}. In addition, we fine-tune pre-trained CLIP \cite{clip} with a lightweight CLIP-Adapter \cite{adp} module to achieve better task adaptation and maintain  generalization. Consequently, our approach combines the advantages of LLM, VLM, and object-centric representation. Our method is simple and efficient. With only $\sim$0.26M trainable parameters, our model can achieve a 97.75\% success rate of placement with seen objects and instructions. Our method generalizes well, achieving an 87.25\% success rate of placement on unseen references and a 77.1\% success rate of placement on unseen instructions. Our method is sample efficient, with only 25\% training data, we still perform better than the top competing approach. Our primary contributions are:

\begin{itemize}

\item We provide a sample efficient object-centric framework, formulating language-conditioned placement as grounding reference objects and spatial relations.
\item We exploit the priors of the pre-trained language model and the visual-language model to generalize better to unseen instructions and objects, and improve the sample efficiency. 
\item We fine-tune the VLM with a lightweight module named CLIP-Adapter, which effectively improves the grounding performance.
\item We evaluate our system in a series of scenarios with seen and unseen objects and language instructions, validating its generalization and effectiveness.

\end{itemize}

\section{Related Works}

\subsection{Foundation Models in Language-Conditioned Manipulation}

As natural language is a user-friendly interface, language-conditioned manipulation has recently become a popular research topic in robotics. In particular, recent advances in foundation models\cite{clip, gpt} have led to considerable progress in learning generalizable manipulation skills under more flexible language instructions. Some works take LLMs as planner for long-horizon tasks \cite{sm, saycan}, take VLMs as part of the agent architecture \cite{cliport, paragon, vilg}, or as part of the reward modeling \cite{minedojo}. Our approach utilizes GPT-3 \cite{gpt} as an instruction parser and fine-tunes CLIP \cite{clip} on our dataset as part of the policy. 

\subsection{Robotic Object Placement}

Several approaches have been proposed for object placement tasks. Some provide goal images \cite{gctn} or collect additional demonstrations \cite{tn}, which are often unscalable and infeasible. Thus, recent works focus more on language-conditioned object placement. Some works focusing on placement \cite{mees1, mees2, semantic} use hand-crafted rules and syntactic structures to parse instructions. Some \cite{clip} learn directly from the raw images. The most recent work \cite{paragon} attempts to combine rule-based and embedding-based methods, but requires large amounts of training data.

In this paper, we focus on the task of language-conditioned object placement. Our approach differs from previous work in that we incorporate the foundation models into an object-centric framework. We use GPT-3 \cite{gpt} to parse instructions into tuples so that we can handle more flexible instructions without increasing the training data. We leverage fine-tuned CLIP\cite{clip} to encode visual and textual object-centric representations. As a whole, we learn faster and generalize better than competing approaches.

\begin{figure*}[t]
    \centering
    \includegraphics[width=0.87 \linewidth]{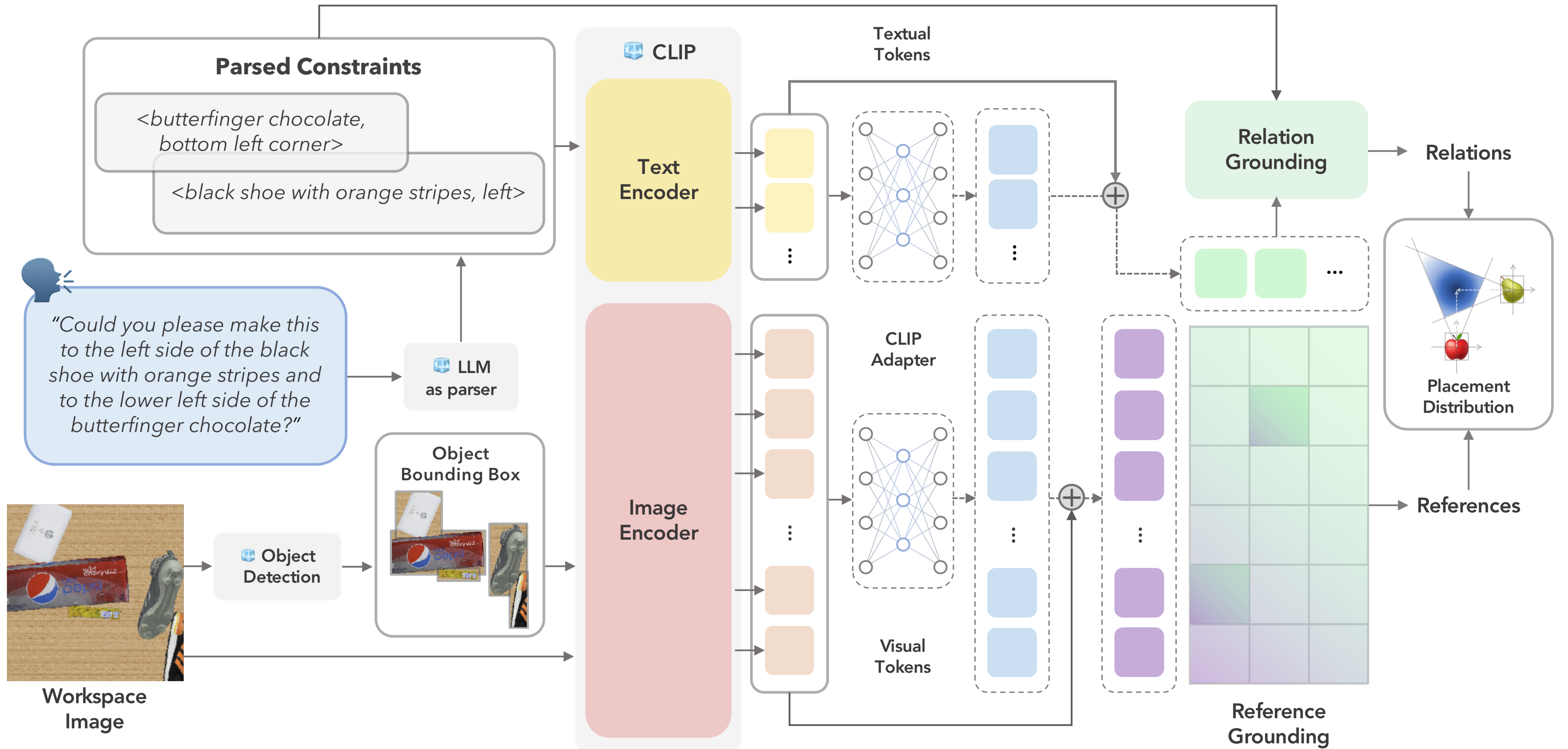}
    \caption{\textbf{System Overview.} Our system parses the language instruction into tuples by GPT-3\cite{gpt} and obtains bounding boxes by detection module(see Sec. \ref{sec:ps}). We then encode textual and visual object-centric representations by fine-tuned CLIP \cite{clip, adp}. Similarities between the adapted textual and visual tokens are computed for reference object grounding. Parsed tuples and textual tokens are used for relation grounding. Placement distribution is then generated based on grounded reference objects and corresponding spatial relations.} \label{fig:ov}

    \vspace{-11pt}
    
\end{figure*} 

\section{Methods}

\subsection{System Overview} 
As shown in Fig. \ref{fig:ov}, we use the pre-trained GPT-3 \cite{gpt} to parse the language instruction $\ell$ and extract language tuples $\Omega=\left\{\left\langle\omega_{\text{ref}}^{l}, \omega_{\text{rel}}^{l}\right\rangle\right\}_{l=1}^{L}$, where $\omega_{\text{ref}}^{l}$ means the reference object name, $\omega_{\text{rel}}^{l}$ means the spatial relation expression, and $L$ is the number of reference-relation pairs contained in the instruction. We utilize the pre-trained CLIP \cite{clip} image and text encoders to encode the workspace RGB image $\mathbf{o}$, the object crops $\left\{a_n\right\}_{1 \leq n \leq N}$ obtained from the detection module, and the language tuples. Given the visual and textual tokens, we propose to use visual and textual CLIP-Adapters \cite{adp} to achieve better task adaptation. Then the cosine similarities between adapted visual and textual tokens are calculated and the best-matching visual token for each textual token is selected as a corresponding reference object token. Meanwhile, if GPT-3 can successfully extract predefined spatial relations, they are applied directly. Otherwise, the corresponding relations of the textual tokens are selected based on their feature similarities to the predefined relations. Finally, we generate placement distribution $p\left(\mathbf{x}\mid \ell, \mathbf{o}\right)$ according to grounded reference-relation pairs using a hand-crafted truncated Gaussian distribution.

\subsection{Object-centric Representation} \label{sec:ps}

\begin{figure}[tb]
    \centering
    \includegraphics[scale=0.17]{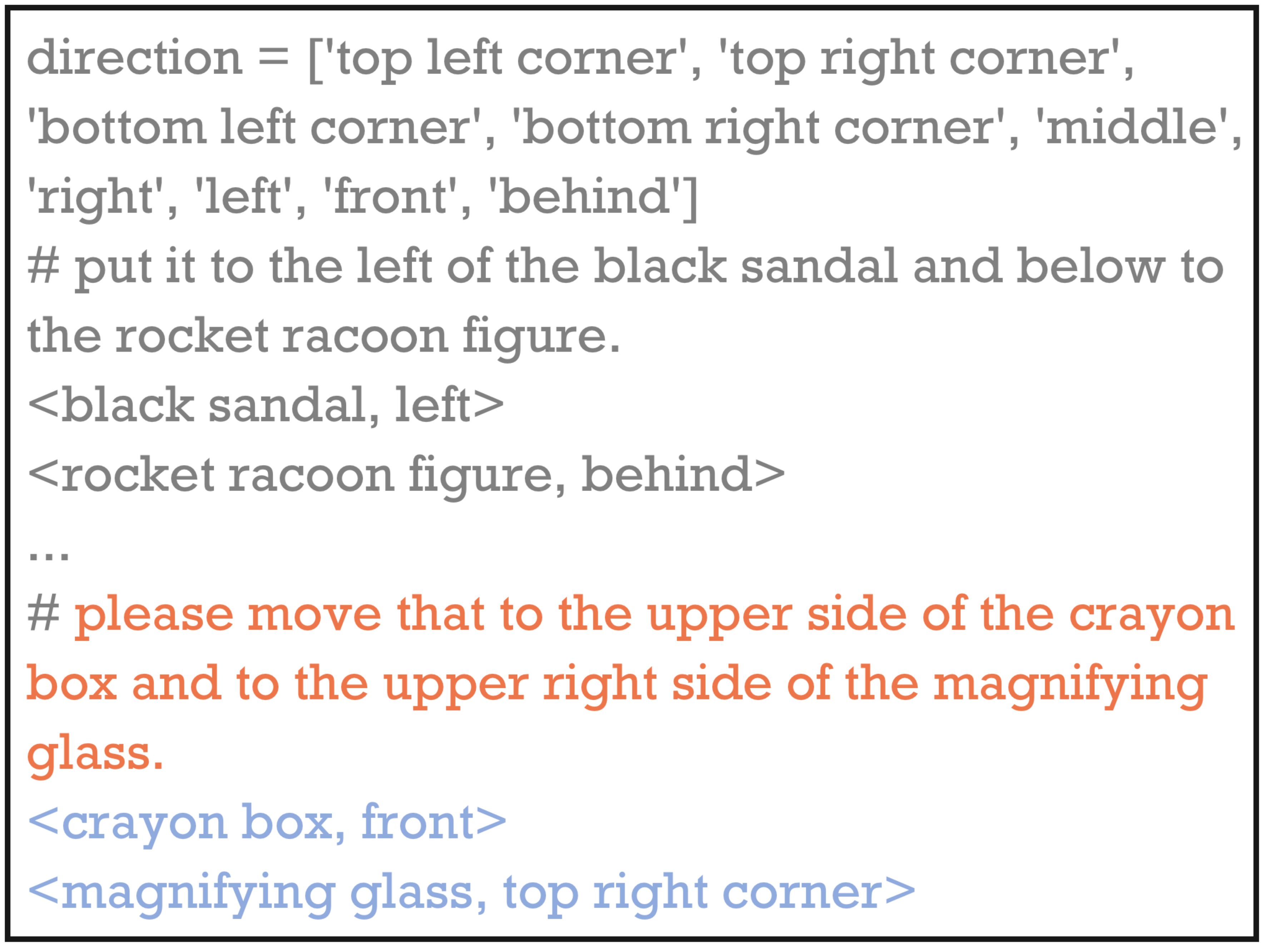}

    \vspace{-10pt}
    
    \caption{Example prompt to parse free-form language instructions into fixed-format tuples. The prompt context is shown in \textcolor{mygrey}{gray}, the input task commands in \textcolor{myorg}{orange}, and the generated outputs in \textcolor{myblue}{blue}.}\label{fig:prompt}

    \vspace{-32pt}
    
\end{figure}

\textbf{Parsing Instructions with LLM.} Similar to Zeng et. al. \cite{sm}, we use the pre-trained GPT-3 \cite{gpt} to parse free-form language instructions into a sequence of fixed-format tuples, {\it i.e.}\textless reference object, spatial relation\textgreater. Such a parsing process turns a flexible instruction into a fixed-format object-centric representation. The power of large language models is derived from large corpora of other data, such as spreadsheets, fictional novels, and questions from standardized tests. We take advantage of the summarizing and writing capabilities of LLMs to generate fixed-format tuples by providing them with several language parsing examples as prompts. An example prompt is shown in Fig. \ref{fig:prompt}. Note that the cases we use in the prompt context are also generated from the ``Seen Instructions'' in TABLE \ref{tab:inst} for fair generalization evaluation. We parse the free-form language instruction $\ell$ into fixed-format tuples, denoted as $\Omega=\left\{\left\langle\omega_{\text{ref}}^{l}, \omega_{\text{rel}}^{l}\right\rangle\right\}_{l=1}^{L}$. 

\textbf{Extracting Object Crops with Detection Module.} Object-centric representations make it easier for robots to understand task scenes. In this paper, object crops are extracted for object-centric representation. For an image consisting of $M$ objects, $N$ bounding boxes are extracted from the detection model ($N \leq M$). $N$ object crops $\left\{a_n\right\}_{1 \leq n \leq N}$ are then obtained by cropping the raw image with the corresponding bounding boxes. Then we feed the object image crops into the pre-trained CLIP \cite{clip} visual encoder. For simplicity, we follow \cite{vilg} to generate object bounding boxes from mask images in Pybullet\cite{pybullet}.

\subsection{Reference Object Grounding with VLM and Adapter} 
\textbf{Pre-processing with Pre-trained VLM.} We propose to use CLIP\cite{clip} to generate visual and textual tokens. Contrastive vision-language pre-training on millions of image-text pairs provides alignment between the visual and textual tokens. We apply CLIP image encoder to encode the N bounding box crops $\left\{a_n\right\}_{1 \leq n \leq N}$ and the workspace image $\mathbf{o}$ and obtain $N+1$ visual tokens $\left\{v_n\right\}_{1 \leq n \leq N+1}$. And the CLIP text encoder consumes the parsed tuples $\left\{\left\langle\omega_{\text {ref}}^{l}, \omega_{\text {rel}}^{l}\right\rangle\right\}_{l=1}^{L}$ and outputs $L$ textual tokens $\left\{t_n\right\}_{1 \leq l \leq L}$. Visual and textual tokens can be stacked into matrices, denoted as $\mathbf{V} \in \mathbb{R}^{(N+1) \times D}$ and $\mathbf{T} \in \mathbb{R}^{L \times D} $, where $D$ represents the feature dimensionality. 

\begin{table*}[]
\caption{Instruction Templates Used in Dataset}
\label{tab:inst}

\vspace{-10pt}

\begin{center}
\begin{tabular}{@{}c|cc|cccc@{}}
\toprule
Tasks & \multicolumn{2}{c|}{Seen Instructions} & \multicolumn{4}{c}{Unseen Instructions} \\ \midrule
Template & \multicolumn{2}{c|}{\begin{tabular}[c]{@{}c@{}}1. put it {[}rel.{]}+{[}ref.{]}.\\ 2. put it {[}rel.{]}+{[}ref.{]}+``and''+{[}rel.{]}+{[}ref.{]}.\end{tabular}} & \multicolumn{4}{c}{\begin{tabular}[c]{@{}c@{}}1. {[}prefix{]}+{[}verb{]}+{[}pron.{]}+{[}rel.{]}+{[}ref.{]}.\\ 2. {[}prefix{]}+{[}verb{]}+{[}pron.{]}+{[}rel.{]}+{[}ref.{]}.+``and''+{[}rel.{]}+{[}ref.{]}.\end{tabular}} \\ \midrule
Components & \multicolumn{1}{c|}{{[}rel.{]} for Table} & {[}rel.{]} for Objects & \multicolumn{1}{c|}{{[}prefix{]}} & \multicolumn{1}{c|}{{[}verb{]}} & \multicolumn{1}{c|}{{[}pron.{]}} & {[}rel.{]}: Spatial Relation Expressions \\ \midrule
\# of Choices & \multicolumn{1}{c|}{9} & 8 & \multicolumn{1}{c|}{7} & \multicolumn{1}{c|}{8} & \multicolumn{1}{c|}{3} & 25 \\ \midrule
Examples & \multicolumn{2}{l|}{``put it to the right part of the table.''} & \multicolumn{4}{l}{\begin{tabular}[c]{@{}l@{}}``could you please drop the thing to the right rear corner of the \\ spiderman figure and on the left hand side of the black shoe\\ with orange and black stripes.''\end{tabular}} \\ \bottomrule
\end{tabular}
\end{center}

\vspace{-21pt}

\end{table*}

\textbf{Fine-Tuning the VLM with Lightweight Adapters.} The ``pretraining-finetuning paradigm'' offers a good solution to better task adaptation. We use CLIP-Adapter \cite{adp} which applies a simple residual transformation layer over the feature tokens generated by CLIP. Thanks to the residual connection and the bottleneck linear layer design, the two 3-layer MLPs significantly improve the grounding performance in both seen and unseen cases. We deploy both a visual adapter $A_v(\cdot)$ and a textual adapter $A_t(\cdot)$ with gating ratios $\alpha, \beta=0.2$ to balance and mix the knowledge from the original tokens and the CLIP-Adapter outputs:

\begin{equation}
    \begin{aligned}
        \mathbf{V}^{\star} = \alpha A_{v}(\mathbf{V}) + (1-\alpha)\mathbf{V} \\
        \mathbf{T}^{\star} = \beta A_{t}(\mathbf{T}) + (1-\beta)\mathbf{T}
    \end{aligned}
\end{equation}

\textbf{Grounding Reference Objects by Similarity.} By calulating the cosine similaity of $\mathbf{V}^{\star}$ and $\mathbf{T}^{\star}$, we can obtain an $L \times (N+1)$-dimensional logits matrix. As in \cite{clip, adp}, we then apply a Softmax function over the second dimension with a temperature parameter to convert it into a probability matrix $\mathbf{P} = \left[p_{l,n}\right]_{L \times (N+1)}  \in \mathbb{R}^{L \times (N+1)}$. To perform reference object grounding, the visual token that has the highest similarity score with a textual token is selected as the match, i.e. $ \hat{n}^{(l)} = \argmax_n p_{l,n}$. This also means that a reference object has been visually grounded for the $l$-th language tuple $\left\langle\omega_{\text{reg}}^{l}, \omega_{\text{rel}}^{l}\right\rangle$. 

\subsection{Placement Distribution Generation}

\textbf{Grounding Relations with LLM.} As stated in Sec. \ref{sec:ps}, GPT-3 \cite{gpt} has the powerful ability to parse language instructions into language tuples. In most cases, it is able to generalize the relational information contained in instructions to the predefined spatial relations, shown as \textcolor{mygrey}{direction} in Fig. \ref{fig:prompt}. In these cases, we use the extracted relations $\omega_{rel}^{l}$ directly. 

However, in a few cases, especially when faced with unseen instructions, GPT-3 may fail to generalize the relational information to the predefined spatial relations. For example, it may generate ``\textless table, rear right corner\textgreater'' instead of ``\textless table, bottom right corner\textgreater''. In these cases, we replace the relation part of the tuple one by one with predefined spatial relations({\it e.g.} replace ``rear right corner'' with ``top left corner'', ``top right corner'', etc.). We then choose the one that is most similar to the unreplaced tuple by calculating cosine similarities between their CLIP textual features. Or, in rare cases where GPT-3 even cannot output a tuple form, we perform the above operation directly with GPT-3's outputs and the predefined relations.

\textbf{Generating Placement Distribution with Grounded Reference-Relation Pairs.}

\begin{figure}[tb]
    \centering
    \includegraphics[scale=0.14]{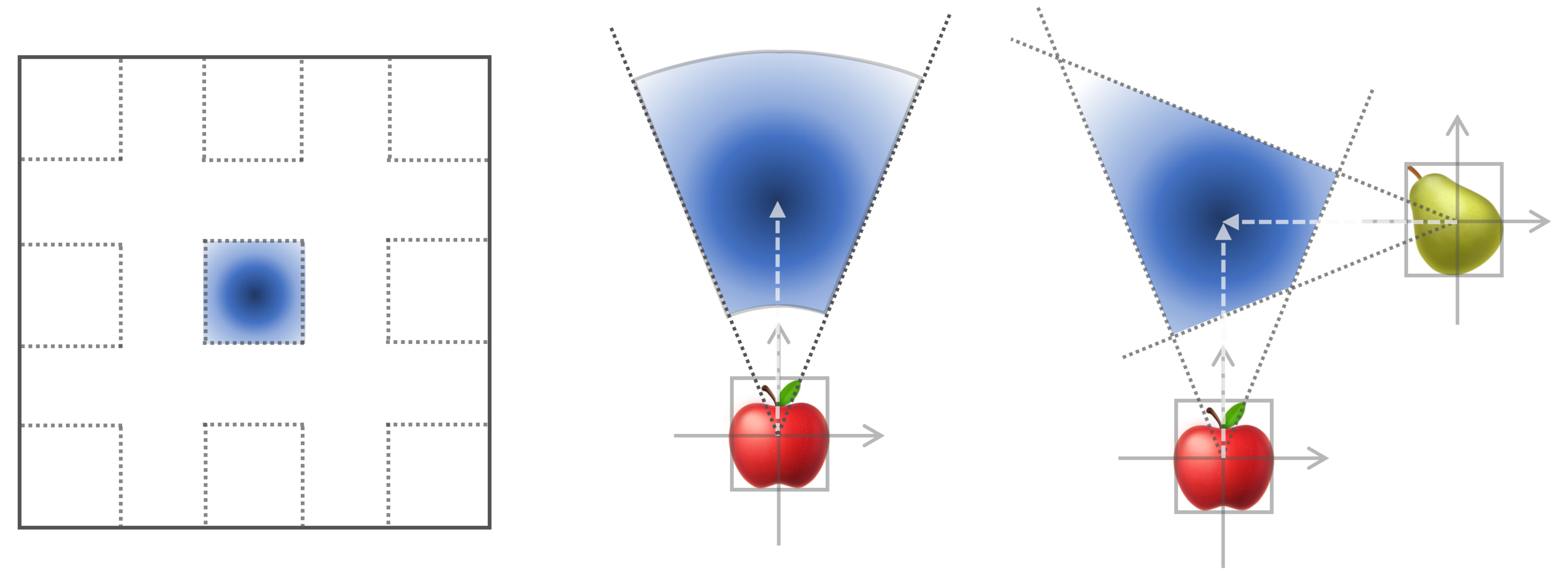}
    
    \vspace{-8pt}
    
    \makebox[0.144\textwidth]{\scriptsize Case (a)}
    \makebox[0.144\textwidth]{\scriptsize Case (b)}
    \makebox[0.144\textwidth]{\scriptsize Case (c)}
    
    \vspace{-8pt}
    \caption{Placement Distribution Generation. The placement is sampled from a truncated Gaussian distribution. Case (a) illustrates the ``to the middle of the table'' distribution, where the darkest region indicates the highest probability of placement. Case (b) illustrates the ``in front of the apple'' distribution. We select the front region and truncated the area that was too close or too far from the reference object. Case (c) illustrates the composition of the spatial relations ``in front of the apple'' and ``left to the pear''. In this case, we take the average sum of the two Gaussian distributions and truncate it with their respective spatial constraints.}\label{fig:dis}

    \vspace{-18pt}
    
\end{figure}

We propose a hand-crafted policy for placement generation. Specifically, we sample from a truncated Gaussian distribution, three examples of which are shown in Fig. \ref{fig:dis}. We also use the axis-aligned bounding boxes (AABB) of all objects to truncate areas where collisions may occur.

Since we are able to ground reference objects and relations with high accuracy, it is appropriate to use hand-crafted placement rules. The placement strategies used in previous work may have collision problems \cite{mees1, mees2}, or may rely on large amounts of training data \cite{paragon}. In contrast, our straightforward placement strategy can produce satisfactory placements and guarantees no collisions.

\subsection{End-to-end Training}
We impose supervision at grounding reference objects instead of placement. The weights of $A_v(\cdot)$ and $A_t(\cdot)$ are optimized with the cross-entropy loss:

\begin{equation}
    \mathcal{L}(\theta)=-\frac{1}{L} \sum_{l=1}^{L} \sum_{n=1}^{N+1} y_{n}^{(l)} \log \hat{y}_{n}^{(l)}
\end{equation}

$y_n^{(l)}=1$ if $n$ equals to the ground-truth category reference object label for the $l$-th tuple $\left\langle\omega_{\text{reg}}^{l}, \omega_{\text{rel}}^{l}\right\rangle$, otherwise $y_n^{(l)}=0$; $\hat{y}_{n}^{(l)}=p_{l,n}$ is the predicted probability for class $n$; $\theta$ represents all the learnable parameters in $A_v(\cdot)$ and $A_t(\cdot)$.

\begin{table*}[t]
\begin{center}
\caption{Results: Success Rate (\%)}
\label{tab:table1}
\begin{tabular}{@{}ccccccccccccc@{}}
\toprule
                            & \multicolumn{4}{c}{seen ref. \& seen inst.}                                                                                & \multicolumn{4}{c}{unseen ref. \& seen inst.}                                                                              & \multicolumn{4}{c}{seen ref. \& unseen inst.}                                                                             \\ \cmidrule(l){2-13} 
\multirow{-2}{*}{Method}    & table                         & 1obj                         & 2obj                         & overall                      & table                         & 1obj                         & 2obj                         & overall                      & table                        & 1obj                         & 2obj                         & overall                      \\ \midrule
CLIPort-Place                     & 31.11                         & 12.22                        & 10.45                        & 15.50                        & 27.78                         & 10.00                        & 7.30                         & 12.52                        & 16.67                        & 14.44                        & 7.73                         & 11.25                        \\
Raw Image Grids             & 45.56                         & 34.44                        & 23.18                        & 30.75                        & 38.89                         & 30.00                        & 20.91                        & 27.00                        & 47.78                        & 22.22                        & 11.36                        & 22.00                        \\
Ours  w/o CLIP pre-train    & 73.33                         & 65.56                        & 47.73                        & 57.50                        & 58.89                         & 33.33                        & 28.18                        & 36.25                        & 73.33                        & 52.22                        & 40.00                        & 50.25                        \\
Ours  w/o GPT-3 (GT-ref \#) & {\color[HTML]{FD6864} 100.00} & {\color[HTML]{FD6864} 97.78} & 63.18                        & 79.25                        & {\color[HTML]{FD6864} 100.00} & 68.00                        & 33.64                        & 56.30                        & 25.56                        & 20.00                        & 9.55                         & 15.50                        \\
Ours w/o Adapter            & 78.89                         & 76.67                        & 65.00                        & 70.75                        & 47.78                         & 78.89                        & {\color[HTML]{FD6864} 82.73} & 74.00                        & 74.44                        & 64.44                        & 44.55                        & 55.75                        \\
Ours                        & {\color[HTML]{FD6864} 100.00} & {\color[HTML]{FD6864} 97.78} & {\color[HTML]{FD6864} 96.82} & {\color[HTML]{FD6864} 97.75} & {\color[HTML]{FD6864} 100.00} & {\color[HTML]{FD6864} 92.22} & 80.00                        & {\color[HTML]{FD6864} 87.25} & {\color[HTML]{FD6864} 93.33} & {\color[HTML]{FD6864} 82.22} & {\color[HTML]{FD6864} 68.36} & {\color[HTML]{FD6864} 77.10} \\ \bottomrule
\end{tabular}
\end{center}

\vspace{-22pt}

\end{table*}

\section{Experiments}
We conduct a series of experiments to evaluate our system. The goals of the experiments are: 1) to demonstrate that our object-centric formulation is effective for language-conditioned placement; 2) to evaluate the generalization performance of our policy on unseen objects and language instructions; 3) to show sample efficiency of our method. 

\subsection{Competing Approaches}

We compare the performance of our system with the following baselines: 

\textbf{CLIPort-Place} is a method that learns placement from the raw image. It is a variant of CLIPort \cite{cliport}, which has a two-stream fully convolutional architecture that uses pre-trained CLIP to ground semantic concepts. Placement generation is learned by predicting the pixel-wise affordance of the raw image. Specifically, we use the attention stream of CLIPort \cite{cliport} for placement. 

\textbf{Raw Image Grids} is a policy that uses CLIP \cite{clip} and CLIP-Adapter \cite{adp} to process raw image grids and parsed tuples. To combine spatial features, 2D positions of the grid centers are projected into a non-linear space (positional embedding as in \cite{nerf}), followed by an MLP. The policy then adds visual features with these positional embeddings. It computes the similarities between the textual tokens and visual tokens and selects the grids with the highest sum of cosine similarities with all the textual tokens. Except for the addition of the positional embeddings, the network architecture is the same as ours. 

We also compare our methods with a series of ablation methods: \textbf{Ours w/o CLIP-Adapter}, \textbf{Ours w/o CLIP pre-train} and \textbf{Ours w/o GPT-3 (GT-Ref\#)}. \textbf{Ours w/o GPT-3 (GT-Ref\#)} is an approach that uses CLIP\cite{clip} and CLIP-Adapter\cite{adp} to process the bounding boxes and the whole instruction sentence. Without LLM's parsing, it's given a ground truth reference-relation pair number $L$. It computes the similarities between the textual token and visual tokens and selects the $L$ highest-scoring reference object(s). Then an MLP consumes the concatenation of the selected visual tokens and the textual token and regresses the corresponding directions.

\subsection{Dataset}

Our dataset is collected in the PyBullet \cite{pybullet} simulation environment with a top-down camera of Intel RealSense L515. Our dataset consists of 2K scenes for training and 1.2K scenes for testing. Each scene contains 5 randomly dropped objects that may be partially occluded, to reflect the challenging language grounding in the real world. The objects are sampled from the Google Scanned Objects dataset \cite{obj}, with the same split between seen and unseen objects as CLIPort \cite{cliport}. Our dataset also contains tables with three seen and three unseen textures. 

We matched each scene to a language instruction generated from predefined templates in TABLE \ref{tab:inst}. The training set contains 450 scenes with a table as a reference, 450 with a single object as a reference, and 1100 with 2 objects as references. The test set consists of three subsets: seen-object-seen-instructions, unseen-object-seen-instructions, and seen-object-unseen-instructions, each with 400 scenes. Each subset has the same ratio of 3 test levels as the training set, {\it i.e.} $9:9:22$. 

\subsection{Evaluation Metrics}
We evaluate the success rate of object placement. Successful placements should satisfy all the spatial constraints in the language instructions and be collision-free. The placement should be in the correct region of the table or not be too close or too far away from the reference objects, with thresholds of $0.1m$ and $0.3m$ respectively. The evaluated models are all trained for 200K steps with a batch size of 1.

\subsection{Results}

\begin{figure}[tb] \centering
    \includegraphics[width=0.98 \linewidth]{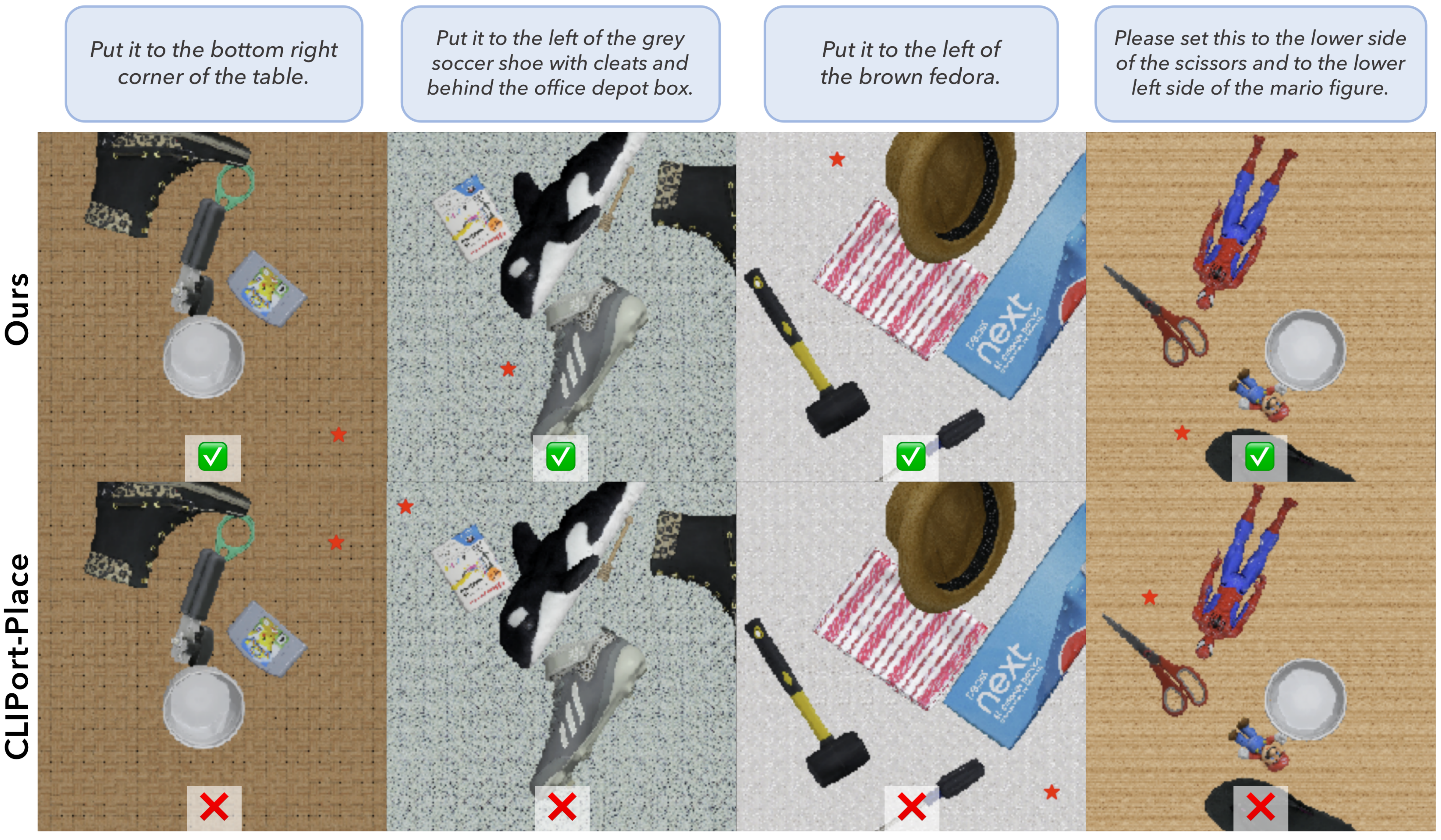}
    \vspace{-9pt}
    
    \caption{Case Visualization. The two on the left are seen cases, with a table and two objects as references. The two on the right are with unseen objects and unseen instructions. The placements marked with stars are sampled from \textbf{Ours} distribution or the $\argmax$ of \textbf{CLIPort-Place} affordance.} \label{fig:case}
    \vspace{-21pt}
    
\end{figure}

\textbf{Effectiveness of Proposed Method.} Table \ref{tab:table1} shows that the effectiveness of our method outperforms all baselines. \textbf{CLIPort-Place} has weak performance in all cases (see Fig. \ref{fig:case}), suggesting that grounding semantics in whole-sentence embeddings from the raw image may not work with only 2000 training samples. In addition, CLIPort \cite{cliport} uses a fully convolutional architecture, which can capture strong local correlations and helps generate ``inside'' placement. However, the experimental results show that it has worse performance when it comes to non-local correlations such as ``left to'' in our task. \textbf{Raw Image Grids} gets better performance than \textbf{CLIPort-Place}. This increase in performance is mainly due to the use of grid crops, which helps to focus on the target zone. However, there is still little object-level information in grids, which may explain its unsatisfactory performance. Instead, with our object-centric formulation, the last 4 methods perform much better. Although limited by the lack of prior knowledge of the pre-trained CLIP, \textbf{Ours w/o CLIP} pre-train still outperforms the formers, with an overall success rate of around 60\%. It suggests that parsing task-relevant textual and visual information, {\it i.e.} tuples and object crops is beneficial for learning. We notice that the performance of \textbf{Ours w/o GPT-3 (GT-Ref\#)} drops a lot from 100\% to 79.25\% when it comes to instructions with 2 reference objects. This may be due to the fact that \textbf{Ours w/o GPT-3 (GT-Ref\#)} fails to extract useful information when the language instruction is long. This may as well explain one aspect of \textbf{CLIPort-Place}'s poor performance -  taking the whole sentence embeddings as input. The longer the sentence, the sparser the valid information, and the worse the performance. \textbf{Ours} has 30\% performance gain over \textbf{Ours w/o Adapter}. It suggests that fine-tuning with CLIP-Adapter is essential to exploit the potential of large models in downstream tasks. 

\textbf{Generalization to Unseen Objects.} Results of cases with 19 unseen objects are also shown in Table \ref{tab:table1}. We observe that our method is capable of generalizing to unseen objects with similar instructions, and achieves the best overall success rate. \textbf{Ours w/o CLIP pre-train} suffers a 20\% performance drop due to the lack of object priors from the pre-trained VLM. And it's worth noting that \textbf{Ours w/o GPT-3 (GT-Ref\#)} achieves 100\% success in table reference cases, but falls by around 30\% percentage in other cases. This indicates that when different object names are contained in instructions, it's hard to focus on the object's textual information without the LLM as a parser. Although \textbf{Ours w/o Adapter} achieves the highest success rate in 2 reference objects cases, it suffers from a decline in performance when seeing tables with less common textures. However, \textbf{Ours} has learned visual features that distinguish a table from other objects. The same applies to \textbf{Ours w/o GPT-3 (GT-Ref\#)}, which also has CLIP-Adapter and achieves 100\% success with the table as a reference. In general, it's glad to see that our method maintains generalization to unseen objects while achieving a high success rate on seen objects. Though, as a trade-off, there may be some overfitting on the seen objects in a few cases. 

\begin{figure}[tb] \centering
    \includegraphics[width=0.47 \linewidth]{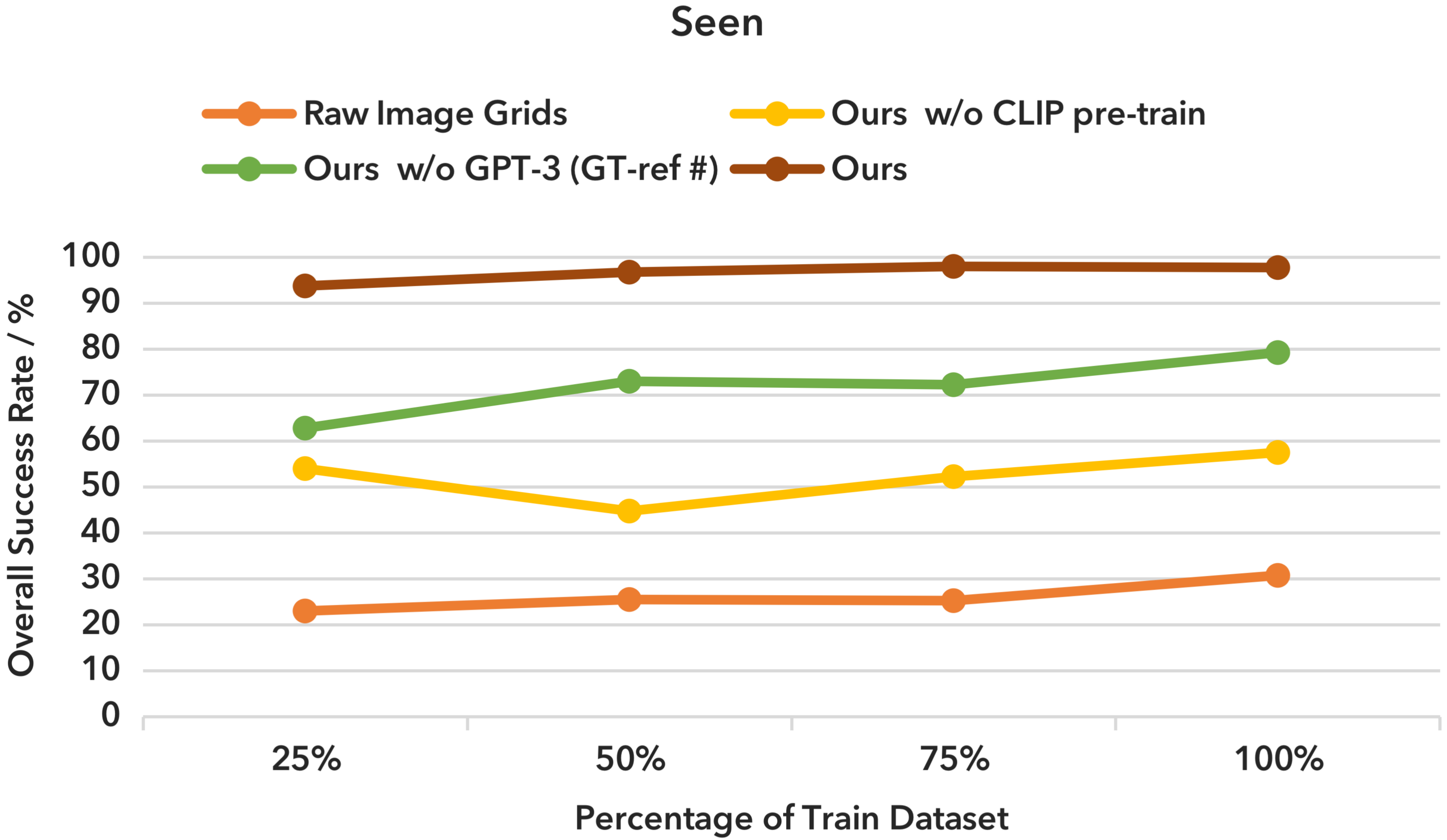}
    \includegraphics[width=0.47 \linewidth]{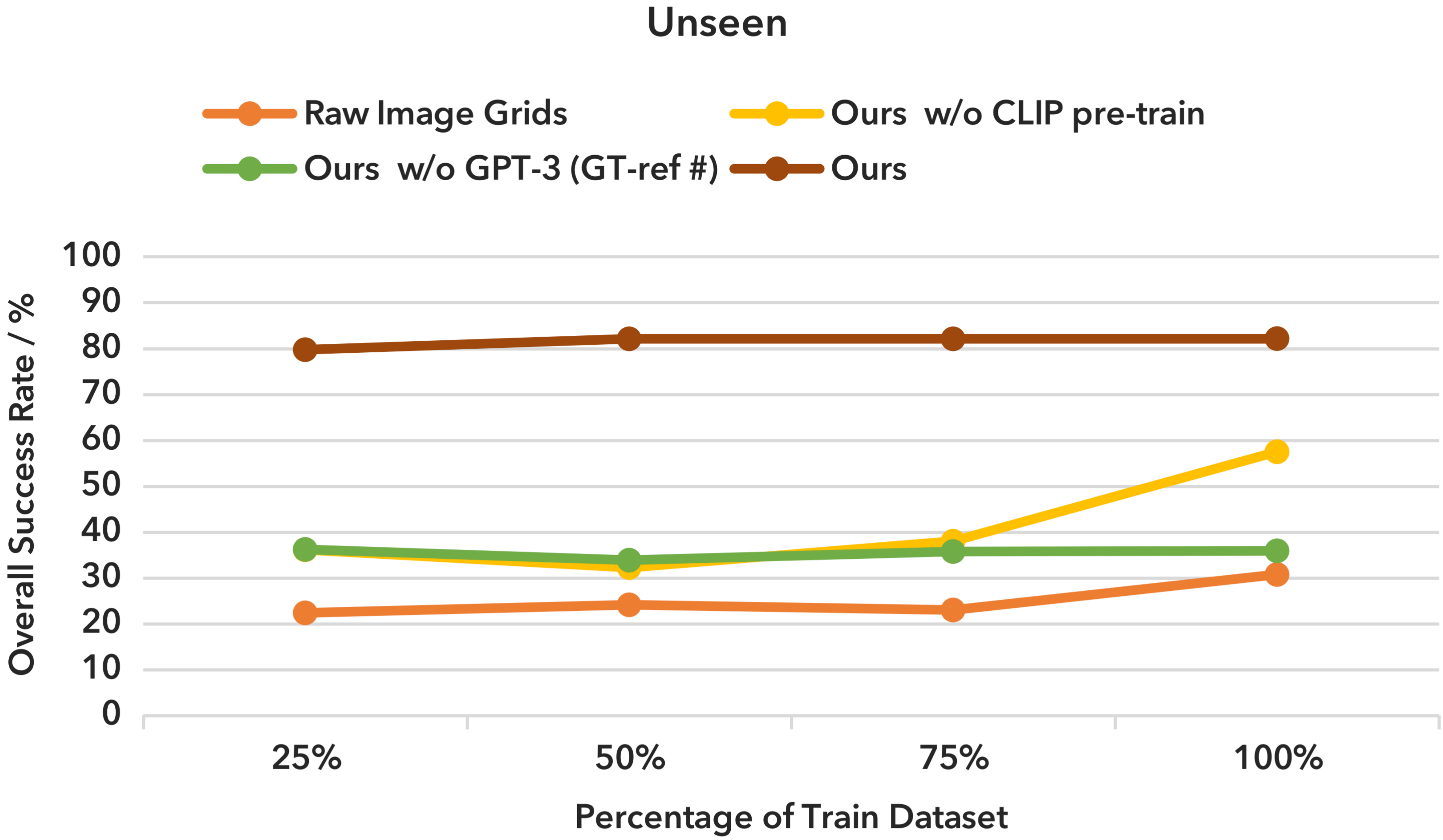}
    
    \vspace{-9pt}
    \caption{Sample Efficiency Evaluation on Seen and Unseen Cases.} \label{fig:dtpic}

    \vspace{-21pt}
    
\end{figure}

\textbf{Generalization to Unseen Instructions.} We test all the models with unseen instructions as well, as shown in TABLE \ref{tab:table1}. With a total of 140,015,400 unseen language instruction options, most models show poor performance. We observe that our method is able to generalize to unseen instructions while all the other models perform considerably worse in the unseen instruction test. It's worth noting that \textbf{Ours w/o GPT-3 (GT-Ref\#)} has the largest drop, showing that language parsing with LLMs is very important to improve the generalization of language instructions.  Since CLIP uses simple language templates for alignment training, the generalization of flexible language instructions may be limited. Thus taking the whole sentence embeddings as input may also account for one aspect of the poor performance of \textbf{CLIPort-Place}. On the contrary, preprocessing with LLM can relieve this problem.

\textbf{Sample Efficiency.} We compare our methods with a series of ablation methods to test: 1) whether our approaches have high sample efficiency; 2) whether our system utilizes samples better than the scene-centric methods, and 3) whether pre-trained CLIP and GPT-3 bring benefit to our policy.

We present the overall success rate versus the percentage of the training dataset in Fig. \ref{fig:dtpic}. The overall success rate for unseen tasks is the $1:1$ average of that with unseen objects and unseen instructions, representing generalization performance. To demonstrate 1), we compare our sample efficiency with \textbf{Raw Image Grids}, \textbf{Ours w/o CLIP pre-train} and \textbf{Ours w/o GPT-3 (GT-Ref\#)}. In general, \textbf{Ours} achieves higher sample efficiency than other alternatives. With only 25\% training data, we still perform better than the top competing approach. For 2), \textbf{Raw Image Grids} has a lower utilization of samples than \textbf{Ours}. This suggests that object-centric representation helps our policy capture visual information, thus improving sample efficiency. To show 3), we compare our policy with 2 ablation methods: \textbf{Ours w/o CLIP pre-train} which trains CLIP\cite{clip} from scratch, and \textbf{Ours w/o GPT-3 (GT-Ref\#)}. It's obvious that without CLIP's prior, \textbf{Ours w/o CLIP pre-train} learns slower than \textbf{Ours} and suffers a performance decline from seen to unseen cases. \textbf{Ours w/o GPT-3 (GT-Ref\#)} shows a performance gain on seen cases when provided more samples, indicating that the model is learning to extract information from a fixed language template. But obviously, without GPT-3's parsing, its performance drops dramatically in terms of novel unseen language instructions.

\section{Conclusions}

In this work, we focus on the task of language-conditioned object placement. We take full advantage of the foundation models. For one thing, we incorporate it into an object-centric framework. For another, we add lightweight adapters to achieve better task adaptation. Consequently, our approach can achieve a higher success rate with less sample even in unseen cases. 

\addtolength{\textheight}{-12cm}  

\bibliographystyle{IEEEtran}
\bibliography{IEEEabrv,ref}
\end{document}